\documentclass[doc,apacite]{apa6}

\usepackage{booktabs} % For formal tables
\usepackage{subcaption}
\usepackage{url}
\usepackage{graphicx}
\usepackage{float}

\usepackage{times}
\usepackage{latexsym}
\usepackage{tabularx}
\usepackage{tablefootnote}

\usepackage{multirow}
\usepackage{makecell}

\usepackage{tikz}
\def\checkmark{\tikz\fill[scale=0.4](0,.35) -- (.25,0) -- (1,.7) -- (.25,.15) -- cycle;}

\title{Towards generalisable hate speech detection: a review on obstacles and solutions}

\shorttitle{Towards generalisable hate speech detection}

\author{Wenjie Yin, Arkaitz Zubiaga}
\affiliation{Queen Mary University of London, London, UK}

\abstract{Hate speech is one type of harmful online content which directly attacks or promotes hate towards a group or an individual member based on their actual or perceived aspects of identity, such as ethnicity, religion, and sexual orientation. With online hate speech on the rise, its automatic detection as a natural language processing task is gaining increasing interest. However, it is only recently that it has been shown that existing models generalise poorly to unseen data. This survey paper attempts to summarise how generalisable existing hate speech detection models are, reason why hate speech models struggle to generalise, sums up existing attempts at addressing the main obstacles, and then proposes directions of future research to improve generalisation in hate speech detection.}

\begin{document}

\maketitle

\section{Introduction}

The Internet saw a growing body of user-generated content as social media platforms flourished \cite{schmidt_survey_2017,chung_conan_2019}. While social media provides a platform for all users to freely express themselves, offensive and harmful content are not rare and can severely impact user experience and even the civility of a community \cite{nobata_abusive_2016}. One type of such harmful content is \textbf{hate speech}, which is speech that \textbf{directly attacks} or \textbf{promotes hate} towards a group or an individual member based on their actual or perceived aspects of \textbf{identity}, such as ethnicity, religion, and sexual orientation \cite{waseem_hateful_2016,davidson_automated_2017,founta_large_2018,sharma_degree_2018}.   % definition explain 

Major social media companies are aware of the harmful nature of hate speech and have policies regarding the moderation of such posts. However, the most commonly used mechanisms are very limited. For example, keyword filters can deal with profanity, but not the nuance in the expression of hate \cite{gao_recognizing_2018}. Crowd-sourcing methods (e.g. human moderators, user reporting), on the other hand, do not scale up. This means that by the time that a hateful post gets detected and taken down, it has already made negative impacts \cite{chen_use_2019}.

The automatic detection of hate speech is thus an urgent and important task. Since the automatic detection of hate speech was formulated as a task in the early 2010s \cite{warner_detecting_2012}, the field has been constantly growing along the perceived importance of the task. 

\subsection{Hate speech, offensive language, and abusive language}% \todo{new}
% abusive and offensive similar but different
% hate speech different from other 
Although different types of abusive and offensive language are closely related, there are important distinctions to note. Offensive language and abusive language are both used as umbrella terms for harmful content in the context of automatic detection studies. However, while ``strongly impolite, rude" and possible use of profanity are seen in the definitions of both \cite{fortuna_survey_2018}, abusive language has a strong component of intentionality \cite{caselli_i_2020}. Thus, offensive language has a broader scope, and hate speech falls in both categories. 

Because of its definition mentioned above, hate speech is also different from other sub-types of offensive language. For example, personal attacks \cite{wulczyn_ex_2017} are characterised by being directed at an individual, which is not necessarily motivated by the target's identity. Hate speech is also different from cyberbullying \cite{zhao2016automatic}, which is carried out repeatedly and over time against vulnerable victims that cannot defend themselves\footnote{for a more elaborate comparison between similar concepts, see \citeA{fortuna_survey_2018}}. This paper focuses on hate speech and generalisation across hate speech datasets, although studies that cover both hate speech and other offensive language are also mentioned.

\subsection{Generalisation}

Most if not all proposed hate speech detection models rely on supervised machine learning methods, where the ultimate purpose is for the model to learn the real relationship between features and predictions through training data, which generalises to previously unobserved inputs \cite{Goodfellow-et-al-2016}. The \textbf{generalisation performance} of a model measures how well it fulfils this purpose.

To approximate a model's generalisation performance, it is usually evaluated on a set-aside test set, assuming that the training and test data, and future possible cases come from the same distribution. 
This is also the main way of evaluating a model's ability to generalise in the field of hate speech detection.

%The assumption that the test set data reflect future unseen data in the same task might not be always true. In the cases where it doesn't, test set performance would not reflect how well the model generalises.

\subsection{Generalisability in hate speech detection}

The ultimate purpose of studying automatic hate speech detection is to facilitate the alleviation of the harms brought by online hate speech.
To fulfil this purpose, hate speech detection models need to be able to deal with the constant growth and evolution of hate speech, regardless of its form, target, and speaker. 

Recent research has raised concerns on the generalisability of existing models \cite{swamy_studying_2019}. Despite their impressive performance on their respective test sets, the performance significantly dropped when the models are applied to a different hate speech dataset. This means that the assumption that test data of existing datasets represent the distribution of future cases is not true, and that the generalisation performance of existing models have been severely overestimated \cite{arango_hate_2020}.
This lack of generalisability undermines the practical value of these hate speech detection models. 
~\\

So far, existing research has mainly focused on demonstrating the lack of generalisability \cite{grondahl_all_2018,swamy_studying_2019,wiegand_detection_2019}, apart from a handful of studies that made individual attempts at addressing aspects of it \cite{waseem_bridging_2018,arango_hate_2020}. Recent survey papers on hate speech and abusive language detection \cite{schmidt_survey_2017, fortuna_survey_2018, al-hassan_detection_2019, mishra_tackling_2019, vidgen_challenges_2019, poletto_resources_2020, vidgen_directions_2020} have focused on the general trends in this field, mainly by comparing features, algorithms and datasets. %\todo{new}
Among these, \citeA{fortuna_survey_2018} provided an in-depth review of definitions, \citeA{vidgen_challenges_2019} concisely summarised various challenges for the detection of abusive language in general, \citeA{poletto_resources_2020} and \citeA{vidgen_directions_2020} created extensive lists of dataset and corpora resources, while \citeA{al-hassan_detection_2019} focused on the special case of the Arabic language.  

This survey paper thus contributes to the literature by providing (1) a comparative summary of existing research that demonstrated the lack of generalisability in hate speech detection models, (2) a systematic analysis of the main obstacles to generalisable hate speech detection and existing attempts to address them, and (3) suggestions for future research to address these obstacles.

\section{Survey Methodology}

%\todo{new}
For each of the three aims of this paper mentioned above, literature search was divided into stages. 

\subsection{Sources of search}

Across different stages, Google Scholar was the main search engine, and two main sets of keywords were used. References and citations were checked back-and-forth, with the number of iterations depending on how coarse or fine-grained the search of that stage was.

\begin{itemize}
    \item General keywords: ``hate speech", ``offensive", ``abusive", ``toxic", ``detection", ``classification".
    \item Generalisation-related keywords: ``generalisation" (``generalization"), ``generalisability" (``generalizability"), ``cross-dataset", ``cross-domain", ``bias".
\end{itemize}

We started with a pre-defined set of keywords. Then, titles of proceedings of the most relevant recent conferences and workshops (Workshop on Abusive Language Online, Workshop on Online Abuse and Harms) were skimmed, to refine the set of keywords. We also modified the keywords during the search stages as we encountered new phrasing of the terms. The above keywords shown are the final keywords.

\subsection{Main literature search stages}

Before starting to address the aims of this paper, an initial coarse literature search involved searching for the general keywords, skimming the titles and abstracts. During this stage, peer-reviewed papers with high number of citations, published in high-impact venues were prioritised. Existing survey papers on hate speech and abusive language detection \cite{schmidt_survey_2017, fortuna_survey_2018, al-hassan_detection_2019, mishra_tackling_2019, vidgen_challenges_2019, poletto_resources_2020, vidgen_directions_2020} were also used as seed papers. The purpose of this stage was to establish a comprehensive high-level view of the current state of hate speech detection and closely related fields.

For the first aim of this paper -- building a comparative summary of existing research on generalisability in hate speech detection -- the search mainly involved different combinations of the general and generalisation-related keywords. As research on this topic is sparse, during this stage, all papers found and deemed relevant were included. 

Building upon the first two stages, the main obstacles towards generalisable hate speech detection were then summarised, as appeared in the section headings of the body of this paper. This was done through extracting and analysing the error analysis of experimental studies found in the first stage, and comparing the results and discussions of the studies found in the second stage. Then, for each category of obstacles identified, another search was carried out, involving combinations of the description and paraphrases of the challenges and the general keywords.
The search in this stage is the most fine-grained, in order to ensure coverage of both the obstacles and existing attempts to address them. %%% include and exclude

After the main search stages, the structure of the main findings in the literature was laid out. During writing, for each type of findings, the most representative studies were included in the writing up. We defined the relative representativeness within studies we have found, based on novelty, experiment design and error analysis, publishing venues, and influence. We also prioritised studies that addressed problems specific to hate speech, compared to better-known problems that are shared with other offensive language and social media tasks.

\section{Generalisation Studies in Hate Speech Detection}\label{subsec:generalisation}

Testing a model on a different dataset from the one which it was trained on is one way to more realistically estimate models' generalisability \cite{wiegand_detection_2019}. 
This evaluation method is called cross-dataset testing \cite{swamy_studying_2019} or cross-application \cite{grondahl_all_2018}, and sometimes cross-domain classification \cite{wiegand_detection_2019} or detection \cite{karan_cross-domain_2018} if datasets of other forms of offensive language are also included. 

As more hate speech and offensive language datasets emerged, %\todo{wording change} 
a number of studies have touched upon cross-dataset generalisation since 2018, either studying generalisability per se, or as part of their model or dataset validation.  These studies are compared in Table \ref{tab:gen_studies}. Most of them carried out cross-dataset testing. %\todo{new}
Genres and publications of the datasets used are summarised in Table \ref{tab:data_gen_studies}. As different datasets and models were investigated, instead of specific performance metrics, the remainder of this section will discuss the general findings of these studies.

\begin{table}[ht]
\centering
%\begin{tabular}{p{1.4cm}|l|p{1.1cm}|p{1.1cm}|p{1.1cm}|p{1.1cm}|p{1.1cm}|p{1.1cm}|p{1.1cm}}\hline
\begin{tabularx}{\textwidth}{p{1.4cm}|p{0.7cm}|p{1.1cm}|p{1.1cm}|p{1.1cm}|p{1.1cm}|p{1.1cm}|p{1.1cm}|p{1.1cm}|p{1.1cm}}
\hline
                          &      & \multicolumn{1}{l}{} & \multicolumn{1}{l}{} & \multicolumn{1}{l}{} & \multicolumn{1}{l}{Study} & \multicolumn{1}{l}{} & \multicolumn{1}{l}{} &  
                          \multicolumn{1}{l}{} \\ 
\cline{3-10}
Dataset name                  & Type &
 \citeA{karan_cross-domain_2018} & \citeA{wiegand_detection_2019}  & \citeA{grondahl_all_2018} & \citeA{waseem_bridging_2018} & \citeA{swamy_studying_2019} & \citeA{arango_hate_2020} & \citeA{fortuna_toxic_2020} & \citeA{caselli_i_2020}  \\ 
\hline
Waseem                   & H*                    & \multicolumn{1}{c|}{\checkmark}  & \multicolumn{1}{c|}{\checkmark}     & \multicolumn{1}{c|}{\checkmark}     & \multicolumn{1}{c|}{\checkmark}   & \multicolumn{1}{c|}{\checkmark}  & \multicolumn{1}{c|}{\checkmark}   & \multicolumn{1}{c|}{\checkmark}    &          \\ 
\hline
Davidson                 & H,O                   &       &          & \multicolumn{1}{c|}{\checkmark}     & \multicolumn{1}{c|}{\checkmark}   & \multicolumn{1}{c|}{\checkmark}  &        & \multicolumn{1}{c|}{\checkmark}    &          \\ 
\hline
Founta                   & H,O                   &       & \multicolumn{1}{c|}{\checkmark}     &          &        & \multicolumn{1}{c|}{\checkmark}  &        &         &          \\ 
\hline
Basile                   & H*                    &       &          &          &        &       & \multicolumn{1}{c|}{\checkmark}   & \multicolumn{1}{c|}{\checkmark}    & \multicolumn{1}{c}{\checkmark}     \\ 
\hline
Kaggle                   & H,O*                  & \multicolumn{1}{c|}{\checkmark}  & \multicolumn{1}{c|}{\checkmark}     &          &        &       &        & \multicolumn{1}{c|}{\checkmark}    &          \\ 
\hline
Gao                      & H                     & \multicolumn{1}{c|}{\checkmark}  &          &          &        &       &        &         &          \\ 
\hline
Fersini                  & H*                    &       &          &          &        &       &        & \multicolumn{1}{c|}{\checkmark}    &          \\ 
\hline
Warner                   & H                     &       & \multicolumn{1}{c|}{\checkmark}     &          &        &       &        &         &          \\ 
\hline
Zhang                    & H                     &       &          & \multicolumn{1}{c|}{\checkmark}     &        &       &        &         &          \\ 
\hline
Kumar                    & O                     & \multicolumn{1}{c|}{\checkmark}  & \multicolumn{1}{c|}{\checkmark}     &          &        &       &        & \multicolumn{1}{c|}{\checkmark}    &          \\ 
\hline
Wulczyn                  & O                     & \multicolumn{1}{c|}{\checkmark}  &          & \multicolumn{1}{c|}{\checkmark}     &        &       &        &         &          \\ 
\hline
Zampieri                  & O                     &       &          &          &        & \multicolumn{1}{c|}{\checkmark}  &        &         & \multicolumn{1}{c}{\checkmark}     \\ 
\hline
Caselli                  & O*                    &       &          &          &        &       &        &         & \multicolumn{1}{c}{\checkmark}     \\ 
\hline
Kolhatkar                & O                     & \multicolumn{1}{c|}{\checkmark}  &          &          &        &       &        &         &          \\ 
\hline
Razavi                   & O                     &       & \multicolumn{1}{c|}{\checkmark}     &          &        &       &        &         &          \\ 
\hline
\multicolumn{2}{l|}{Model}                       & SVM   & FastText & Mixed    & MLP    & BERT  & Mixed  & N/A     & BERT     \\
\hline
\end{tabularx}
\caption{Comparison of studies that looked at cross-dataset generalisation, by datasets and models used. Dataset types: H: hate speech, O: other offensive language, *: contains subtypes. \protect\citeA{fortuna_toxic_2020} compared datasets through class vector representations, the rest of the studies carried out cross-dataset testing. \protect\citeA{grondahl_all_2018} didn't control model type across testing conditions. }
\label{tab:gen_studies}
\end{table}

% \todo{not sure about the order of these two tables}
\begin{table}[htp]
\centering
\begin{tabularx}{\textwidth}{p{1.2cm}|p{3.5cm}|p{2.8cm}|p{2.8cm}|p{1.9cm}} 
\hline
\multicolumn{1}{c|}{\textbf{\makecell{Dataset \\ name}}}  & \multicolumn{1}{c|}{\textbf{\makecell{~ \\ Publication}}} & \multicolumn{1}{c|}{\textbf{\makecell{~ \\ Source}}}                           & \multicolumn{1}{c|}{\textbf{\makecell{~ \\ Positive labels}}}                                                                           & \multicolumn{1}{c}{\textbf{\makecell{Annotator \\ type}}}       \\ 
\hline
\hline
\makecell{Waseem \\ ~}    &  \makecell{\citeA{waseem_hateful_2016} \\ \citeA{waseem_are_2016}}           & \makecell{Twitter \\ ~}                          & \makecell{Racism \\ Sexism}                                                                            & Expert               \\
\hline
Davidson  &  \citeA{davidson_automated_2017}           & Twitter                          & \makecell{Hate speech \\ Offensive}                                                                    & Crowdsourcing        \\
\hline
Founta    &   \citeA{founta_large_2018}          & Twitter                          & \makecell{Hate speech \\ Offensive}                                                                    & Crowdsourcing        \\
\hline
Basile    &   \citeA{basile_semeval-2019_2019}          & Twitter                          & Hateful                                                                                   & Crowdsourcing        \\
\hline
Kaggle    &   \citeA{jigsaw_2018}          & Wikipedia                        & \makecell{Toxic \\ Severe toxic \\ Obscene \\ Threat \\ Insult \\ Identity hate}                               & Crowdsourcing        \\
\hline
Gao       &   \citeA{gao_detecting_2018}          & Fox News                         & Hateful                                                                                   & ? (Native speakers)  \\
\hline
Fersini   &   \citeA{fersini_detecting_2019}          & Twitter                          & Misogynous                                                                                & Expert               \\
\hline
 Warner    &   \citeA{warner_detecting_2012}          & \makecell{Yahoo! \\ American Jewish \\ \hspace{0.5cm} Congress} & \makecell{Anti-semitic \\ Anti-black \\ Anti-asian \\ Anti-woman \\ Anti-muslim \\ Anti-immigrant \\ Other-hate} & ? (Volunteer)        \\
\hline
 Zhang     &    \citeA{zhang_detecting_2018}         & Twitter                          & Hate                                                                                      & Expert               \\
\hline
Kumar     &    \citeA{kumar_aggression-annotated_2018}         & Facebook, Twitter                & \makecell{Overtly aggressive \\ Covertly aggressive}                                                    & Expert               \\
\hline
Wulczyn   &    \citeA{wulczyn_ex_2017}         & Wikipedia                         & Attacking                                                                                 & Crowdsourcing        \\
\hline
Zampieri  &    \citeA{zampieri_predicting_2019}         & Twitter                          & Offensive                                                                                 & Crowdsourcing        \\
\hline
Caselli   &     \citeA{caselli_i_2020}        & Twitter                          & \makecell{Explicit (abuse) \\ Implicit (abuse)}                                                        & Expert               \\
\hline
Kolhatkar &     \citeA{kolhatkar_sfu_2019}        & The Globe and Mail               & \makecell{Very toxic \\ Toxic \\ Mildly toxic}                                                           & Crowdsourcing        \\
\hline
Razavi    &     \citeA{hutchison_offensive_2010}        & \makecell{Natural Semantic \\ \hspace{0.5cm} Module \\ Usenet}  & Flame                                                                                     & Expert               \\
\hline
\end{tabularx}
\caption{Datasets used in cross-dataset generalisation studies. Positive labels are listed with their original wording. Expert annotation type include authors and experts in social science and related fields. ?: Type of annotations not available in original paper, the found descriptions are thus included. Note that only datasets used in generalisation studies are listed -- for comprehensive lists of hate speech datasets, see \protect\citeA{vidgen_directions_2020} and \protect\citeA{poletto_resources_2020}.}
\label{tab:data_gen_studies}
\end{table}

Firstly, \textbf{existing ``state-of-the-art" models had been severely over-estimated} \cite{arango_hate_2020}. 

\citeA{grondahl_all_2018} trained a range of models, and cross-applied them on four datasets %\cite{wulczyn_ex_2017,davidson_automated_2017,waseem_hateful_2016,zhang_detecting_2018}
\textit{(Wulczyn, Davidson, Waseem, Zhang)}. The models included LSTM, which is one of the most popular neural network types in text classification, and  CNN-GRU \cite{zhang_detecting_2018}, which outperformed previous models on six datasets.    
On a different testing dataset, both models' performance dropped by more than 30 points in macro-averaged F1 across the Twitter hate speech datasets. 

More recently, \citeA{arango_hate_2020} also found performance drops of around 30 points in macro-averaged F1 with BiLSTM \cite{agrawal2018deep} and GBDT over LSTM-extracted embeddings \cite{badjatiya2017deep} models when applied on %a different Twitter dataset, ``HatEval" \cite{basile_semeval-2019_2019}
\textit{Basile}. These two models were both considered state-of-the-art when trained and evaluated on %the Twitter dataset by Waseem and Hovy \citeA{waseem_hateful_2016}
\textit{Waseem}. They demonstrated methodological flaws in each: overfitting induced by extracting features on the combination of training and test set; oversampling before cross-validation boosted F1 scores mathematically. \citeA{grondahl_all_2018} also reported that they failed to reproduce \citeA{badjatiya2017deep}'s results.

The most recent popular approach of fine-tuning BERT \cite{bert/corr/abs-1810-04805} is no exception, although the drop is slightly smaller. In a cross-dataset evaluation with four datasets (\textit{Waseem, Davidson, Founta, Zampieri})%\cite{waseem_hateful_2016,davidson_automated_2017,founta_large_2018,zampieri_semeval-2019_2019}
, performance drop ranged from 2 to 30 points in macro-averaged F1 \cite{swamy_studying_2019}.

Similar results were also shown in traditional machine learning models, including character n-gram Logistic Regression \cite{grondahl_all_2018}, character n-gram Multi-Layer Perceptron (MLP) \cite{grondahl_all_2018,waseem_bridging_2018}, linear Support Vector Machines \cite{karan_cross-domain_2018}. The same was true for shallow networks with pre-trained embeddings, such as MLP with Byte-Pair Encoding (BPE)-based subword embeddings \cite{bpe-embeddings2017,waseem_bridging_2018} and FastText  \cite{joulin2017bag, wiegand_detection_2019}. 

\textbf{Generalisation also depends on the datasets that the model was trained and tested on}.

The lack of generalisation highlights the differences in the distribution of posts between datasets \cite{karan_cross-domain_2018}.
While the size of the difference varies, some general patterns can be found. 
Some datasets are more similar than others, as there are groups of datasets that produce models that generalise much better on each other. For example, in \citeA{wiegand_detection_2019}'s study, FastText models \cite{joulin2017bag} trained on three datasets 
\textit{(Kaggle, Founta, Razavi)} achieved F1 scores above 70 when tested on one another, while models trained or tested on datasets outside this group achieved around 60 or less. The authors \cite{wiegand_detection_2019} attributed this to the higher percentage of explicit abuse in the samples and less biased sampling procedures. In \citeA{swamy_studying_2019}'s study with fine-tuned BERT models \cite{bert/corr/abs-1810-04805}, 
\textit{Founta} and \textit{Zampieri}
produced models that performed well on each other, which was considered an effect of the similar characteristics shared between these two datasets, given that similar search terms were used for building the datasets.

So far, there has been only one study that attempted to quantify the similarity between datasets. \citeA{fortuna_toxic_2020} used averaged word embeddings \cite{fasttext/corr/BojanowskiGJM16,mikolov_advances_2017} to compute the representations of classes from different datasets, and compared classes across datasets. 
One of their observations is that \textit{Davidson}
's ``hate speech" is very different from \textit{Waseem}
's ``hate speech",``racism",``sexism", while being relatively close to 
\textit{Basile}'s ``hate speech" and \textit{Kaggle}
's ``identity hate". This echoes with experiments that showed poor generalisation of models from 
\textit{Waseem}
to \textit{Basile}
\cite{arango_hate_2020} 
and between
\textit{Davidson} 
and \textit{Waseem} 
\cite{waseem_bridging_2018,grondahl_all_2018}. 

Training on some datasets might produce more generalisable models, but in terms of which datasets or what properties of a dataset lead to more generalisable models, there is not enough consistency. \citeA{swamy_studying_2019} holds that a larger proportion of abusive posts (including hateful and offensive) leads to better generalisation to dissimilar datasets, such as \textit{Davidson}. This is in line with \citeA{karan_cross-domain_2018}'s study where \textit{Kumar} and 
\textit{Kolhatkar} 
generalised best, and \citeA{waseem_bridging_2018}'s study where models trained on  \textit{Davidson} 
generalised better to \textit{Waseem} 
than the other way round. In contrast, \citeA{wiegand_detection_2019} concluded that the proportion of explicit posts and less biased sampling played the most important roles: \textit{Kaggle} and \textit{Founta} 
generalised best, despite being the datasets with the least abusive posts. %\todo{new}
\citeA{caselli_i_2020} found that, on \textit{Basile}, a BERT \cite{bert/corr/abs-1810-04805} model trained on the dataset they proposed (\textit{Caselli}) outperformed the one trained on \textit{Basile} end-to-end. They attributed this to their quality of annotation as well as to a bigger data size. This is encouraging, yet more synthesis across different studies is needed surrounding this very recent dataset.

\section{Obstacles to Generalisable Hate Speech Detection}\label{subsec:obstacles}

Demonstrating the lack of generalisability is only the first step in understanding this problem. 
This section delves into three key factors that may have contributed to it: (1) presence of non-standard grammar and vocabulary, (2) paucity of and biases in datasets, and (3) implicit expressions of hate.

\subsection{Non-standard Grammar and Vocabulary on Social Media}\label{subsec:linguistic}

On social media, non-standard English is widely used. 
This is sometimes shown in a more casual use of syntax, such as the omission of punctuation \cite{blodgett_racial_2017}. 
Alternative spelling and expressions are also used in dialects \cite{blodgett_racial_2017}, to save space, and to provide emotional emphasis \cite{baziotis_datastories_2017}.

% \todo{rearranged}
Hate speech detection, which is largely focused on social media, shares the above challenges and has its specific ones.
%shares this challenge with many other Natural Language Processing (NLP) tasks on social media posts, and is also particularly affected by it.
Commonly seen in hate speech, the offender adopts various approaches to evade content moderation. For example, the spelling of offensive words or phrases can be obfuscated \cite{nobata_abusive_2016,serra_class-based_2017}, %\todo{moved from implicit section}
and common words such as ``Skype", ``Google", and ``banana" may have a hateful meaning -- sometimes known as euphemism or code words \cite{taylor_surfacing_2017, magu_determining_2018}. 
 
These unique linguistic phenomena pose extra challenge on training generalisable models, mainly by making it difficult to utilise common NLP pre-training approaches. %\todo{new}
When the spelling is obfuscated, a word is considered out-of-vocabulary and thus no useful information can be given by the pre-trained models.
In the case of code words, pre-trained embeddings will not reflect its context-dependent hateful meaning. At the same time, simply using identified code words for a lexicon-based detection approach will result in low precision \cite{davidson_automated_2017}.
As there are infinite ways of combining the above alternative rules of spelling, code words, and syntax, hate speech detection models struggle with these rare expressions even with the aid of pre-trained word embeddings.

In practice, this difficulty is manifested in false negatives. \citeA{qian_leveraging_2018} found that rare words and implicit expressions are the two main causes of false negatives; \citeA{aken_challenges_2018} compared several models that used pre-trained word embeddings, and found that rare and unknown words were present in 30\% of the false negatives of Wikipedia data and 43\% of Twitter data. Others have also identified rare and unknown words as a challenge for hate speech detection \cite{nobata_abusive_2016,zhang_hate_2018}.

\textbf{Existing solutions}

% \todo{moved from implicit section, extended}
From a domain-specific perspective, \citeA{taylor_surfacing_2017} and \citeA{magu_determining_2018} attempted to \textbf{identify code words} for slurs used in hate communities. Both of them used keyword search as part of their sourcing of Twitter data and word embedding models to model word relationships. %, but they used very different approaches to identify hateful communities and discover code words. 
\citeA{taylor_surfacing_2017} identified hate communities through Twitter connections of the authors of extremist articles and hate speech keyword searches. They trained their own dependency2vec \cite{levy2014dependency} and FastText \cite{fasttext/corr/BojanowskiGJM16} embeddings on the hate community tweets and randomly sampled ``clean" tweets, and used weighted graphs to measure similarity and relatedness of words. Strong and weak links were thus drawn from unknown words to hate speech words. In contrast, \citeA{magu_determining_2018} collected potentially hateful tweets using a set of known code words. They then computed the cosine similarity between all words based on word2vec \cite{mikolov2013distributed} pre-trained on news data. Code words, which have a neutral meaning in news context, were further apart from other words which fit in the hate speech context. Both studies focused on the discovery of such code words and expanding relevant lexicons, but their methods could potentially complement existing hate lexicons as classifier features or for data collection. 

Recently, a lot more studies approached the problem by adapting well-known embedding methods to hate speech detection models.

The benefit of \textbf{character-level features} has not been consistently observed. 
%•	Attempts at char-level models
Three studies compared character-level, word-level, and hybrid (both character and word-level) CNNs, but drew completely different conclusions. \citeA{park_finding_2018} and \citeA{meyer_platform_2019} found hybrid and character CNN to perform best respectively. Probably most surprisingly, \citeA{lee_comparative_2018} observed that word and hybrid CNNs outperformed character CNN to similar extents, with all CNNs worse than character n-gram logistic regression.
Small differences between these studies could have contributed to this inconsistency. More importantly, unlike the word components of the models, which were initialised with pre-trained word embeddings, the character embeddings were trained end-to-end on the very limited respective training datasets. It is thus likely that these character embeddings severely overfit on the training data. 

In contrast, simple character n-gram logistic regression has shown results as good as sophisticated neural network models, including the above CNNs \cite{aken_challenges_2018,gao_detecting_2018,lee_comparative_2018}. Indeed, models with fewer parameters are less likely to overfit. This suggest that character-level features themselves are very useful, when used appropriately.
A few studies used word embeddings that were additionally enriched with subword information as part of the pre-training. For example, FastText \cite{fasttext/corr/BojanowskiGJM16} models were consistently better than hybrid CNNs \cite{bodapati_neural_2019}. MIMICK \cite{mimick/corr/PinterGE17}-based model displayed similar performances \cite{mishra_neural_2018}. 

The use of \textbf{sentence embeddings} partially solves the out-of-vocabulary problem by using the information of the whole post instead of individual words. Universal Sentence Encoder \cite{USE/abs-1803-11175}, combined with shallow classifiers, helped one team \cite{indurthi_fermi_2019} achieve first place at HatEval 2019 \cite{basile_semeval-2019_2019}. Sentence embeddings, especially those trained with multiple tasks, also consistently outperformed traditional word embeddings \cite{chen_use_2019}.

\textbf{Large language models} with sub-word information have the benefits of both subword-level word embeddings and sentence embeddings. They produce the embedding of each word with its context and word form. Indeed, BERT \cite{bert/corr/abs-1810-04805} and its variants, have demonstrated top performances at hate or abusive speech challenges recently \cite{liu_nuli_2019,mishra_3idiots_2019}. 
 %wording... idk

Nonetheless, these relatively good solutions to out-of-vocabulary words (subword- and context-enriched embeddings) all face the same short-coming: they have only seen the standard English in BookCorpus and Wikipedia. 
NLP tools perform best when trained and applied in specific domains \cite{duarte_mixed_2018}. In hate speech detection, word embeddings trained on relevant data (social media or news sites) had a clear advantage \cite{chen_comparison_2018,vidgen_detecting_2020}. The domain mismatch could have similarly impaired the subword- and context-enrich models' performances.% but research on the equivalent of these models pre-trained on social media is still absent \cite{bodapati_neural_2019}. 

%In short, hate speech detection as a task that involves predominantly social media data faces non-standard language use and out-of-vocabulary words in the data. End-to-end character-level models are ineffective due to overfitting, pre-trained subword- or context-enriched embeddings are of different domains than social media.

\subsection{Limited, Biased Labelled Data}

\subsubsection{Small data size}
%Hate speech detection is a supervised task, which requires labelled data. 
It is particularly challenging to acquire labelled data for hate speech detection %For some tasks, labelled data can be crawled from existing platforms. For example, sentiment analysis can utilise text of movie or product reviews and the stars/scores users gave. In contrast, not only do hate speech data need to be manually labelled, 
as knowledge or relevant training is required of the annotators. As a high-level and abstract concept, the judgement of ``hate speech" is subjective, needing extra care when processing annotations. Hence, datasets are usually not big in size. 

When using machine learning models, especially deep learning models with millions of parameters, small dataset size can lead to overfitting and harm generalisability \cite{Goodfellow-et-al-2016}. 

\textbf{Existing solutions}

The use of \textbf{pre-trained embeddings} (discussed earlier) and parameter dropout \cite{srivastava2014dropout} have been accepted as standard practice in the field of NLP to prevent over-fitting, and are common in hate speech detection as well. Nonetheless, the effectiveness of domain-general embedding models is questionable, and there has been only a limited number of studies that looked into the \textit{relative} suitability of different pre-trained embeddings on hate speech detection tasks \cite{chen_comparison_2018,mishra_neural_2018,bodapati_neural_2019}. 

% \todo{new}
In \citeA{swamy_studying_2019}'s study of model generalisability, \textbf{abusive language-specific pre-trained embeddings} were suggested as a possible solution to limited dataset sizes. \citeA{alatawi_detecting_2020} proposed White Supremacy Word2Vec (WSW2V), which was trained on one million tweets sourced through white supremacy-related hashtags and users. Compared to general word2vec \cite{mikolov2013distributed} and GloVe \cite{pennington_glove_2014} models trained on news, Wikipedia, and Twitter data, WSW2V captured meaning more suitable in the hate speech context -- e.g. ambiguous words like ``race" and ``black" have higher similarity to words related to ethnicity rather than sports or colours. Nonetheless, their WSW2V-based LSTM model did not consistently outperform Twitter GloVe-based LSTM model or BERT \cite{bert/corr/abs-1810-04805}. %They did not consider cross-dataset testing for generalisablity, either.

% transfer learning
% \todo{rearranged}
Research on \textbf{transfer learning from other tasks}, such as sentiment analysis, also lacks consistency.
% \todo{rearranged}
\citeA{uban_transfer_2019} pre-trained a classification model on a large sentiment  dataset\footnote{https://help.sentiment140.com/}, and performed transfer learning on the \textit{Zampieri} and \textit{Kumar} datasets. 
They took pre-training further than the embedding layer, comparing word2vec \cite{mikolov2013distributed} to sentiment embeddings and entire-model transfer learning.
Entire-model transfer learning was always better than using the baseline word2vec \cite{mikolov2013distributed} model, but the transfer learning performances with only the sentiment embeddings were not consistent. 

% \todo{new}
More recently, \citeA{cao_deephate_2020} also trained sentiment embeddings through classification as part of their proposed model. The main differences are: the training data was much smaller, containing only \textit{Davidson} and \textit{Founta} datasets; the sentiment labels were produced by VADER \cite{gilbert2014vader}; their model was deeper and used general word embeddings \cite{mikolov2013distributed, pennington_glove_2014, wieting2015paraphrase} and topic representation computed through Latent Dirichlet Allocation (LDA) \cite{blei2003latent} in parallel. Through ablation studies, they showed that sentiment embeddings were beneficial for both \textit{Davidson} and \textit{Founta} datasets.

Use of existing knowledge from a more mature research field like that of sentiment analysis has the potential to be used to jumpstart relatively newer fields, but more investigation into the conditions in which transfer learning works best has to be done.

\subsubsection{Sampling bias}

Non-random sampling makes datasets prone to bias. Hate speech and, more generally, offensive language generally represent less than 3\% of social media content \cite{zampieri_semeval-2019_2019,founta_large_2018}.
To alleviate the effect of scarce positive cases on model training, all existing social media hate speech or offensive content datasets used boosted (or focused) sampling with simple heuristics. 

Table \ref{tab:datasets_sampling} compares the \textbf{sampling methods} of hate speech datasets studied the most in cross-dataset generalisation. Consistently, keyword search and identifying potential hateful users are the most common methods. However, what is used as the keywords (slurs, neutral words, profanity, hashtags), which users to include (any user from keyword search, identified haters), and the use of other sampling methods (identifying victims, sentiment classification) all vary a lot.

\begin{table}[htb]
\centering
\begin{tabularx}{\textwidth}{p{1.3cm}|p{5.5cm}|p{3cm}|p{3cm}} 
\hline
Dataset  & Keywords                                                                                   & Haters                           & Other                               \\ 
\hline
Waseem   & ``Common slurs and terms used pertaining to religious, sexual, gender, and ethnic minorities" & ``A small number of prolific users" & N/A                                 \\ 
\hline
Davidson & HateBase\tablefootnote{https://www.hatebase.org/}                                                                                   & ``Each user from lexicon search"    & N/A                                 \\ 
\hline
Founta   & HateBase, NoSwearing\tablefootnote{https://www.noswearing.com/dictionary/}                                                                       & N/A                              & Negative sentiment                  \\ 
\hline
Basile   & ``Neutral keywords and derogatory words against the targets, 
highly polarized hashtags"      & ``Identified haters"                & ``Potential victims
of hate accounts"  \\
\hline
\end{tabularx}
\caption{Boosted sampling methods of the most commonly studied hate speech datasets \protect\cite{waseem_hateful_2016,davidson_automated_2017,founta_large_2018,basile_semeval-2019_2019}. Description as appeared in
the publications. N/A: no relevant descriptions found.} 
\label{tab:datasets_sampling}
\end{table}

\begin{table}[p]
\centering
\begin{tabularx}{\textwidth}{p{1.3cm}|p{4cm}|p{3cm}|p{4cm}} 
\hline
Dataset  & Action                                                                                                                                                                                                                       & Target                                                                                                                           & Clarifications                                                                                                                                                                                                                  \\ 
\hline
Waseem   & \underline{Attacks, seeks to silence,} \underline{criticises}, \textit{negatively stereotypes, promotes hate speech or violent crime, blatantly misrepresents truth or seeks to distort views on}, uses a sexist or racial slur, defends xenophobia or sexism & A minority                                                                                                                       & (Inclusion) Contains a screen name that is offensive, as per the previous criteria, the tweet is ambiguous
(at best), and the tweet is on a topic that satisfies any of the above criteria                                      \\ 
\hline
Davidson & \textit{Express hatred towards}, \underline{humiliate, insult}*                                                                                                                                                                                   & A group or members of the group                                                                                                  & (Exclusion) Think not just about the words appearing in a given tweet but about the context in which they were used; the presence of a particular word, however offensive, did not necessarily indicate a tweet is hate speech  \\ 
\hline
Founta   & \textit{Express hatred towards}, \underline{humiliate, insult}                                                                                                                                                                                    & Individual or group, on the basis of attributes
such as race, religion, ethnic origin, sexual orientation,
disability, or gender & N/A                                                                                                                                                                                                                             \\ 
\hline
Basile   & \textit{Spread, incite, promote, justify hatred or violence towards}, \underline{dehumanizing,} \underline{hurting or intimidating}**                                                                                                                           & Women or immigrants                                                                                                              & (Exclusion) Hate speech against other targets, offensive language, blasphemy, historical denial, overt incitement to terrorism, offense towards public servants and police officers, defamation                                 \\
\hline
\end{tabularx}
\caption{Annotation guidelines of the most commonly studied hate speech datasets. Original wording from the publications or supplementary materials; action verbs grouped for easier comparison: \underline{underlined:} directly attack or attempt to hurt, \textit{italic:} promote hate towards. N/A: no relevant descriptions found. *\protect\citeA{davidson_automated_2017} also gave annotators ``a paragraph explaining it (the definition) in further detail", which was not provided in their publication.
**\protect\citeA{basile_semeval-2019_2019} also gave annotators some examples in their introduction of the task (rather than the main guidelines, thus not included).}
\label{tab:datasets_annotation}
\end{table}

Moreover, different studies are based on varying definitions of ``hate speech", as seen in different \textbf{annotation guidelines} (Table \ref{tab:datasets_annotation}). Despite all covering the same two main aspects (directly attack or promote hate towards), datasets vary by their wording, what they consider a target (any group, minority groups, specific minority groups), and their clarifications on edge cases.  %\todo{extended} 
\textit{Davidson} and \textit{Basile} both distinguished ``hate speech" from ``offensive language", while ``uses a sexist or racist slur" is in \textit{Waseem}'s guidelines to mark a case positive of hate, blurring the boundary of offensive and hateful. Additionally, as both \textit{Basile} and \textit{Waseem} specified the types of hate (towards women and immigrants; racism and sexism), hate speech that fell outside of these specific types were not included in the positive classes, while \textit{Founta} and \textit{Davidson} included any type of hate speech.  Guidelines also differ in how detailed they are: Apart from \textit{Founta}, all other datasets started the annotation process with sets of labels pre-defined by the authors, among which \textit{Waseem} gave the most specific description of actions. In contrast, \textit{Founta} only provided annotators with short conceptual definitions of a range of possible labels, allowing more freedom for a first exploratory round of annotation. After that, labels were finalised, and another round of annotation was carried out. As a result, the labelling reflects how the general public, without much domain knowledge, would classify offensive language. For example, the ``abusive" class and ``offensive" class were so similar that they were merged in the second stage. However, as discussed above, they differ by whether intentionality is present \cite{caselli_i_2020}.

Such different annotation and labelling criteria result in essentially different tasks and different training objectives, despite their data having a lot in common. 

As a result of the varying and sampling methods, definitions, and annotation schemes, what current models can learn on one dataset is specific to the examples in that dataset and the task defined by the dataset, limiting the models' ability to generalise to new data. 

One type of possible resulting bias is author bias. For example, 
65\% of the hate speech in the \textit{Waseem} dataset was produced by merely two users, and their tweets exist in both the training and the test set. Models trained on such data thus overfit to these users' language styles. This overfitting to authors was proven in two state-of-the-art models \cite{badjatiya2017deep,agrawal2018deep} \cite{arango_hate_2020}.
Topic bias is another concern. With words such as  ``football" and ``announcer" among the ones with the highest Pointwise Mutual Information (PMI) with hate speech posts, a topic bias towards sports was demonstrated in the \textit{Waseem} dataset \cite{wiegand_detection_2019}. 

\textbf{Existing solutions}

A few recent studies have attempted to go beyond one dataset when training a model. \citeA{waseem_bridging_2018} used \textbf{multitask training} \cite{caruana1997multitask} with hard parameter sharing up to the final classification components, which were each tuned to one hate speech dataset. The shared shallower layers, intuitively, extract features useful for both datasets, with the two classification tasks as regularisation against overfitting to either one.
Their multitask-trained models matched the performances of models trained end-to-end to single datasets and had clear advantage over simple dataset concatenation, whilst allowing generalisation to another dataset. \citeA{karan_cross-domain_2018} presented a similar study. Frustratingly Easy \textbf{Domain Adaptation} \cite{daume2007frustratingly}, %which essentially duplicates features to account for different domains, 
had similar beneficial effects but was much simpler and more efficient. These two studies showed the potential of combining datasets to increase generalisability, but further investigation into this approach is lacking. 

% transfer learning from similar tasks..

\subsubsection{Representation bias}

%\todo{added reasoning}
Natural language is a proxy of human behaviour, thus the biases of our society are reflected in the datasets and models we build. With increasing real-life applications of NLP systems, these biases can be translated into wider social impacts \cite{hovy_social_2016}.
Minority groups are underrepresented in available data and/or data annotators, thus causing biases against them when models are trained from this data. This phenomenon is also seen in audio transcribing \cite{tatman2017gender}, sentiment analysis \cite{kiritchenko2018examining}, etc.

Hate speech detection models not only have higher tendency to classify African-American English posts as offensive or hate than ``white" English \cite{davidson_racial_2019}, but also more often predict false negatives on ``white" than African-American English \cite{sap_social_2019}.
Certain words and phrases, including neutral identity terms such as ``gay" \cite{dixon_measuring_2018} and ``woman" \cite{park_reducing_2018} can also easily lead to a false positive judgement. %\todo{new}
Moreover, just like biases in real life, racial, gender, and party identification biases in hate speech datasets were found to be intersectional \cite{kim_intersectional_2020}. Unlike the other types of biases mentioned above, rather than performance metrics such as the overall F1 score, they do more harm to the practical value of the automatic hate speech detection models. These biases may cause automatic models to amplify the harm against minority groups instead of mitigating such harm as intended \cite{davidson_racial_2019}. For example, with higher false positive rates for minority groups, their already under-represented voice will be more often falsely censored. 

\textbf{Existing solutions}

%\todo{pretty much completely rewritten... not sure about the level of detail}
Systematic studies of representation biases and their mitigation are relatively recent. Since \citeA{dixon_measuring_2018} first quantified unintended biases in abusive language detection on the \textit{Wulczyn} dataset using a synthetic test set, an increasing number of studies have been carried out on hate speech and other offensive language.
These attempts to address biases against minority social groups differ by how they measure biases and their approaches to mitigate them.

Similar to \citeA{dixon_measuring_2018}, a number of studies measured bias as certain words and phrases being associated with the hateful or offensive class, which were mostly identity phrases.  Attempts to mitigate biases identified this way focus on decoupling this association between features and classes. 
Model performance on a \textbf{synthetic test set} with classes and identity terms balanced, compared to the original test data, were used a measure for model bias. 
Well-known identity terms and synonyms are usually used as starting points \cite{dixon_measuring_2018, park_reducing_2018, nozza_unintended_2019}. Alternatively, bias-prone terms could be identified through looking at skewed distributions within a specific dataset \cite{badjatiya_stereotypical_2019, mozafari_hate_2020}.

A few studies measured biases across directly \textbf{predicted language styles or demographic attributes} of authors. \citeA{davidson_racial_2019} and \citeA{kim_intersectional_2020} both tested their hate speech detection models on  \citeA{blodgett2016demographic}'s distantly supervised dataset of African-American vs white-aligned English tweets, revealing higher tendencies of labelling an African-American-aligned tweet offensive or hateful. \citeA{kim_intersectional_2020} further extended this observation to gender and party identification. As the testing datasets do not have hateful or offensive ground truth labels, one caveat is that, using this as a metric of model bias assumes that all language styles have equal chances of being hateful or offensive, which might not be true. 

\citeA{huang_multilingual_2020} approached author demographics from a different angle, and instead predicted author demographics on available hate speech datasets using user profile descriptions, names, and photos. They built and released a multilingual corpus for model bias evaluation.
Although now with ground truth hate speech labels, this introduces additional possible bias existing in the tools they used into the bias evaluation process. For example, they used a computer vision API on the profile pictures to predict race, age, and gender, which displayed racial and gender biases \cite{buolamwini2018gender}. 
%As explicit features like keywords are not used and the predictions themselves are not guaranteed to be reflecting the demographics of the authors, mitigation is even less straightforward.  

One mitigation approach that stemmed from the first approach of measuring biases is ``debiasing" training data through \textbf{data augmentation}. \citeA{dixon_measuring_2018} retrieved non-toxic examples containing a range of identity terms following a template, which were added to \textit{Wulczyn}. Following a similar logic, \citeA{park_reducing_2018} created examples containing the counterpart of gendered terms found in the data to address gender bias in the \textit{Waseem} and \textit{Founta} datasets. \citeA{badjatiya_stereotypical_2019} extended this word replacement method by experimenting with various strategies including named entity tags, part of speech tags, hypernyms, and similar words from word embeddings, which were then applied on the \textit{Wulczyn} and \textit{Davidson} datasets.

Less biased \textbf{external corpora and pre-trained models} could also be used.  To reduce gender bias, \citeA{park_reducing_2018} also compared pre-trained debiased word embeddings \cite{bolukbasi_man_2016} and transfer learning from a larger, less biased corpus. Similarly, \citeA{nozza_unintended_2019} added samples from the \textit{Waseem} dataset to their training dataset (\textit{Fersini}), to keep classes and gender identity terms balanced.

From the perspective of model training, ``debiasing" could also be integrated into the \textbf{model training objective}. Based on 2-grams' Local Mutual Information with a label, \citeA{mozafari_hate_2020} gave each training example in the \textit{Davidson} and \textit{Waseem} datasets a positive weight, producing a new weighted loss function to optimise. \citeA{kennedy_contextualizing_2020} built upon a recent study of post-hoc BERT feature importance \cite{jin_towards_2019}. A regularisation term to encourage the importance of a set of identity terms to be close to zero was added to the loss function. This changed the ranks of importance beyond the curated set of identity terms in the final model trained on two datasets \cite{de_gibert_hate_2018, kennedy_gab_2018}, with that of most identity terms decreasing, and some aggressive words increasing, such as ``destroys", ``poisoned". \citeA{vaidya_empirical_2019} used a similar multitask learning framework to \citeA{waseem_bridging_2018} on \textit{Kaggle}, but with the classification of author's identity as the auxiliary task to mitigate the confusion between identity keywords and hateful reference. %Their multitask learning model performed better than their baseline, but only when self-attention is also in place. Thus, it is not clear whether multitask learning itself is making the difference, or how much the self-attention contributed. 

There is little consensus in how bias and the effect of bias mitigation should be measured, with different studies adopting varying ``debiased" metrics, including Error Rate Equality Difference \cite{dixon_measuring_2018, park_reducing_2018, nozza_unintended_2019}, pinned AUC Equality Difference \cite{dixon_measuring_2018, badjatiya_stereotypical_2019}, Pinned Bias \cite{badjatiya_stereotypical_2019}, synthetic test set AUC \cite{park_reducing_2018}, and weighted average of subgroup AUCs \cite{nozza_unintended_2019, vaidya_empirical_2019}. 
More importantly, such metrics are all defined based on how the subgroups are defined -- which datasets are used, which social groups are compared, which keywords or predictive models are chosen to categorise those groups. %These studies were also on different datasets and different types of biases. 
As a consequence, although such metrics provide quantitative comparison between different mitigation strategies within a study, the results are hard to compare horizontally. 
Nonetheless, a common pattern is found across the studies: the standard metric, such as raw F1 or AUC, and the ``debiased" metrics seldom improve at the same time.
This raises the question on the relative importance that should be put on ``debiased" metrics and widely accepted raw metrics: How much practical value do such debiased metrics have if they contradict raw metrics? Or do we need to rethink the widely accepted AUC and F1 scores on benchmark datasets because they do not reflect the toll on minority groups? 

% \todo{rephrased}
In comparison, \citeA{sap_risk_2019} proposed to address the biases of human annotators during dataset building, rather than debiasing already annotated data or regularising models. 
By including each tweet's dialect and providing \textbf{extra annotation instructions} to think of tweet dialect as a proxy of the author's ethnic identity, they managed to significantly reduce the likelihood of the largely white annotator group (75\%) to rate an African-American English tweet offensive to anyone or to themselves. This approach bears similarity to \citeA{vaidya_empirical_2019}'s, which also sought to distinguish identity judgement from offensiveness spotting, although in automatic models.
Although on a small scale, this study demonstrated that more care can be put into annotator instructions than existing datasets have. 
% future studies: models less biased?

\subsection{Hate Expression Can Be Implicit}

Slurs and profanity are common in hate speech. This is partly why keywords are widely used as a proxy to identify hate speech in existing datasets. However, hate can also be expressed through stereotypes \cite{sap_social_2019}, sarcasm, irony, humour, and metaphor \cite{mishra_tackling_2019, vidgen_challenges_2019}.
For example, a post that reads ``Hey Brianne - get in the kitchen and make me a samich. Chop Chop" \cite{gao_detecting_2018} \textit{directly attacks} a woman \textit{based on} her female \textit{identity} using stereotypes, and thus certainly fulfills the definition of hate speech, without any distinctive keyword.  

Implicit hate speech conveys the same desire to distance such social groups as explicit hate speech \cite{alorainy_enemy_2019} and are no less harmful \cite{breitfeller_finding_2019}. % because there is no lexical features to be learnt \cite{wiegand_detection_2019}. 
Implicit expressions are the most commonly mentioned cause of false negatives in error analysis \cite{zhang_hate_2018,qian_leveraging_2018,basile_semeval-2019_2019,mozafari_bert-based_2020}.
% false negatives are harmful because....(cite). (further impact)
Inability to detect nuanced, implicit expressions of hate means the models do not go beyond lexical features and cannot capture the underlying hateful intent, let alone generalise to hate speech cases where there are no recurring hate-related words and phrases.
Because of the reliance on lexical features, automatic detection models fall far short of human's ability to detect hate and are thus far from being applicable in the real world as a moderation tool \cite{duarte_mixed_2018}.

 It has been proposed that abusive language should be systematically classified into explicit and implicit, as well as generalised and directed \cite{waseem_understanding_2017}. Several subsequent studies have also identified nuanced, implicit expression as a particularly important challenge in hate speech detection for future research to address \cite{aken_challenges_2018,duarte_mixed_2018,swamy_studying_2019}. It is especially necessary for explainability \cite{mishra_tackling_2019}.
 Despite the wide recognition of the problem, there has been much fewer attempts at addressing it.

\textbf{Existing solutions}

%Existing attempts to address the problem of implicit hate speech tend to each focus on one particular form of implicit hate speech.
Implicit cases of hate speech are hard to identify because they can be understood only within their specific context or with the help of relevant real-world knowledge such as stereotypes. 
Some have thus \textbf{included context in datasets}. %
For example, \citeA{gao_detecting_2018} included the original news articles as the context of the comments. \citeA{de_gibert_hate_2018}'s hate speech forum dataset organised sentences in the same post together, and has a ``relation" label separate from ``hate"/``no hate" to set apart cases which can only be correctly understood with its neighbours.

% \todo{added new dataset, rearranged}
Offensive or abusive language datasets that include implicitness in annotation schemes have appeared only recently. The \textit{Caselli} dataset \cite{caselli_i_2020} is so far the only \textbf{dataset with a standalone ``implicit" label}. They re-annotated the \textit{Zampieri} dataset \cite{zampieri_predicting_2019}, splitting the offensive class into implicitly abusive, explicitly abusive, and non-abusive. Their dataset thus offered a clearer distinction between abusiveness and offensiveness, and between implicit and explicit abuse.
\citeA{sap_social_2019} asked annotators to explicitly \textbf{paraphrase the implied statements} of intentionally offensive posts. The task defined by this dataset is thus very different from previously existing ones -- it is a sequence-to-sequence task to generate implied statements on top of the classification task to identify hateful intent. 

% \todo{added new dataset, rearranged}
Both of their experiments reveal that predicting implicit abuse or biases remains a major challenge. \citeA{sap_social_2019}'s model tended to output the most generic bias of each social group, rather than the implied bias in each post. \citeA{caselli_i_2020}'s best model achieved only a precision of around .234 and a recall of 0.098 for the implicit class, in contrast to .864 and .936 for non-abusive and .640 and .509 for explicit.

% \todo{new}
To the best of our knowledge, so far there has only been one attempt at annotating the implicitness of hate speech specifically. \citeA{alatawi_detecting_2020} crowd-sourced annotation on a small set of tweets collected through white supremacist hashtags and user names, dividing them into implicit white supremacism, explicit white supremacism, other hate, and neutral. Unfortunately, the inter-annotator agreement was so low (0.11 Cohen's kappa \cite{cohen1960coefficient}) that they reduced the labels into binary (hateful vs non-hateful) in the end. The main disagreements are between neutral and implicit labels. Compared to \citeA{sap_social_2019} and \citeA{caselli_i_2020}'s studies, their result highlights the difficulty of annotating implicit hate speech and, more fundamentally, the perception of hate speech largely depends on the reader, as posited by \citeA{waseem_are_2016}. 

Fewer studies proposed \textbf{model design motivated by implicit hate speech}. \citeA{gao_recognizing_2018} designed a novel two-path model, aiming to capture both explicit hate speech with a ``slur learner" path and implicit hate speech with an LSTM path. However, it is doubtful whether the LSTM path really learns to identify implicit hate speech, as it is also trained on hate speech cases acquired through initial slur-matching and the slur learner. 
 
Targeting specific types of implicit hate speech seems more effective.  \citeA{alorainy_enemy_2019} developed a feature set using dependency trees, part-of-speech tags, and pronouns, to capture the us vs them sentiment in implicit hate speech. This improved classification performance on a range of classifiers including CNN-GRU and LSTM. The main shortcoming is that the performance gain was relative to unprocessed training data, so it is not clear how effective this feature set is compared to common pre-processing methods. 

\section{Discussion}

%\todo{minor changes}
While cross-dataset testing highlights the low generalisability of existing models, it is important to not reduce the study of generalisability in hate speech detection to cross-dataset performance or ``debiased" metrics.
Ultimately, we want generalisability to the real world. 
Why we are developing these models and datasets, how we intend to use them, and what potential impacts they may have on the users and the wider society are all worth keeping in mind. While mathematical metrics offer quantification, our focus should always be on what we plan to address and its context. Furthermore, hate speech datasets and models should be representative of what hate speech is with no prioritising of any facets of it \cite{swamy_studying_2019}, and shouldn't discriminate against minority groups that they are intended to protect \cite{davidson_racial_2019}. 

Hate speech detection as a sub-field of NLP is rather new. Despite the help of established NLP methods, achieving consensus in the formulation of the problem is still ongoing work -- whether it is binary, multi-class, hierarchical, how to source representative data, what metadata should be included, and where we draw the line between offensive and hateful content. Thus, no existing dataset qualifies as a ``benchmark dataset" yet \cite{swamy_studying_2019}. In the near future, it is likely that new datasets will continue to emerge and shape our understanding of how to study hate speech computationally. Thus, while it is important to try to solve the problems defined by existing datasets, more emphasis should be put on generalisability.

\subsection{Future research}
More work can be done from the perspectives of both models and datasets to make automatic hate speech detection generalisable and thus practical.
Here, we lay out critical things to keep in mind for any researcher working on hate speech detection as well as research directions to evaluate and improve generalisability.

\subsubsection{Datasets}

\textbf{Clear label definitions} 

A prerequisite is to have clear label definitions, separating hate speech from other types of offensive language \cite{davidson_automated_2017, founta_large_2018}, and abusive language from offensive language \cite{caselli_i_2020}. In addition to this, to address the ambiguity between types of abusive language, future datasets can cover a wider spectrum of abusive language such as personal attacks, trolling, and cyberbullying. 
 This could be done either in a hierarchical manner like what \citeA{basile_semeval-2019_2019} and \citeA{kumar_aggression-annotated_2018} did with subtypes of hate speech and aggression respectively, or in a multi-label manner, as there might be cases where more than one can apply, as seen in \citeA{waseem_hateful_2016}'s racism and sexism labels. At the same time, the definitions of labels should have as little overlap as possible.

\textbf{Annotation quality}

\citeA{poletto_resources_2020} found that only about two thirds of the existing datasets report inter-annotator agreement. Guidelines also range from brief descriptions of each class to long paragraphs of definitions and examples.
To ensure a high inter-annotator agreement, extensive instructions and the use of expert annotators is required. There exists a trade-off between having a larger dataset and having annotations with a high inter-annotator agreement that reflect an understanding of the concepts. At the same time, extra guidelines were shown to be effective in addressing some of the biases in crowd-sourced annotations \cite{sap_risk_2019}. Future research can look into what type of and how much training or instruction is required to match the annotations of crowdworkers and experts. %Datasets that are bigger in size but with a trade-off on guideline clarity might be less suitable for evaluation of models than being part of the training data.
% Annotators also need to be made aware of implicit expressions and the language styles of various communities.

\textbf{Understanding perception}

The perception of hate speech depends on the background of the person reading it \cite{waseem_are_2016}.
Existing datasets mostly reported the number of annotators and whether they are crowdworkers, but seldom the demographics of annotators. %To reduce annotators' biases, a balance in annotator demographics is needed.
Furthermore, within the range of ``expert" annotators, there are also many possibilities, such as the authors themselves \cite{de_gibert_hate_2018, mandl_overview_2019}, experts in linguistics \cite{kumar_benchmarking_2018}, activists \cite{waseem_are_2016, waseem_hateful_2016}, experts in politics \cite{vidgen_detecting_2020}.  Future studies can investigate what factors contribute to the disagreement between annotators, quantitatively or qualitatively. Datasets with extensive annotator attributes and their judgements could be built. Annotating implicit hate speech is especially challenging \cite{alatawi_detecting_2020}. Through improved understanding of hate speech perception, an implicit hate speech dataset could be made possible.

\textbf{Drawing representative samples}

As discussed above, before the annotation process, how the initial pool of posts is are collected and how the proportion of positive cases is boosted could introduce bias into the dataset. It is a better approach to start with an initial random sample and then apply boosting techniques, compared to drawing a filtered sample \cite{wiegand_detection_2019}.
Boosting techniques can also be improved, by shifting away from keywords towards other less lexical proxies of possible hate. Future datasets should also actively address different types of possible biases, such as regularising each user's contribution to one dataset, analysis of the topics present in the dataset, limiting the association between certain terms or language styles and a label. 

\subsubsection{Models}
    %\item Using multiple hate speech datasets during training. More work can be done on comparing different methods, and what characteristics of the datasets interact with the effectiveness.
    %\item Transfer learning from bigger datasets.  Hate speech detection may benefit from more similar yet developed fields, such as sentiment analysis. Hate speech or abusive language-specific embeddings is also a possibility.
    %\item Reducing the reliance on lexical features. Domain knowledge such as linguistic patterns and underlying sentiment of hate speech can inform model design, feature extraction or preprocessing. How to effectively combine different types of features?

\textbf{Reducing overfitting} 

Overfitting can be reduced through training on more than one dataset \cite{waseem_bridging_2018, karan_cross-domain_2018} or transfer learning from a larger dataset \cite{uban_transfer_2019, alatawi_detecting_2020} and/or a closely related task, such as sentiment analysis \cite{uban_transfer_2019, cao_deephate_2020}, yet synthesis in the literature is lacking. More work can be done on comparing different training approaches, and what characteristics of the datasets interact with the effectiveness. For example, when performing transfer learning, the trade-off between domain-specificity and dataset size and representativeness is worth investigating.

Reducing the reliance on lexical features can also help alleviate overfitting to the training dataset. Domain knowledge such as linguistic patterns and underlying sentiment of hate speech can inform model design, feature extraction or preprocessing \cite{alorainy_enemy_2019}. Future studies can look into how features of different nature can be effectively combined.

\textbf{Debiasing models}

A range of approaches could be used to make the model less biased against certain terms or language styles, from the perspectives of training data or objective. Each study shows that their approach takes some effect, yet comparison across studies is still difficult. More systematic comparisons between debiasing approaches is favourable. This can be done by applying a range of existing approaches on a number of datasets, with a set of consistent definitions of attributes. There could also be an interaction between debiasing approaches and the types of biases. When experimenting with ``debiasing", it is important to always stay critical of any metrics used.

\textbf{Model application and impact}
 
When evaluating models, dataset-wise mathematical metrics like F1/AUC should not be the only measurement. It is also important to evaluate models also on datasets not seen during training \cite{wiegand_detection_2019}, and carry out in-depth error analysis relevant to any specific challenge that the model claims to address. 

Machine learning models should be considered as part of a sociotechnical system, instead of an algorithm which only exists in relation to the input and outcomes \cite{selbst_fairness_2019}. Thus, more future work can be put into studying hate speech detection models in a wider context of application. For example, can automatic models practically aid human moderators in content moderation? In that case, how can human moderators make use of the outputs or post-hoc feature analysis most effectively? Would that introduce more bias or reduce bias in content moderation? What would the impact be on the users of the platform? To answer these questions, interdisciplinary collaboration is needed.

\section{Conclusion}

Existing hate speech detection models generalise poorly on new, unseen datasets. Reasons why generalisable hate speech detection is hard come from limits of existing NLP methods, dataset building, and the nature of online hate speech, and are often intertwined. The behaviour of social media users and especially haters  poses extra challenge to established NLP methods. Small datasets make deep learning models prone to overfitting, and biases in datasets transfer to models. While some biases come from different sampling methods or definitions, others merely reflect long-standing biases in our society. Hate speech evolves with time and context, and thus has a lot of variation in expression. Existing attempts to address these challenges span across adapting state-of-the-art in other NLP tasks, refining data collection and annotation, and drawing inspirations from domain knowledge of hate speech.
More work can be done in these directions to increase generalisability. At the same time, the task shouldn't be framed entirely as an algorithmic one. Instead, wider context and impact should also be considered.

% \section*{Acknowledgments}

% yes

\bibliography{sample,hate_speech}

\end{document}